\documentclass{article}
\usepackage[utf8]{inputenc}
\usepackage[
backend=biber,
style=authoryear-comp,
sorting=ynt
]{biblatex}
\usepackage{amsmath}
\usepackage{amsfonts}
\usepackage[a4paper, total={6in, 8in}]{geometry}
\usepackage{array}
\usepackage{graphicx}
\usepackage{caption}
\usepackage{float}
\usepackage{hyperref}
\usepackage{graphicx}
\graphicspath{ {./images/} }
\title{Hard Label Black Box Node Injection Attack on Graph Neural Network}
\author{Yu Zhou, Zihao Dong, Guofeng Zhang, Jingchen Tang\\
    University of California, Los Angeles \\}
\date{June 15th 2022}

\begin{document}

\maketitle

\begin{abstract}
    While graph neural networks have achieved state-of-the-art performances in many real-world tasks including graph classification and node classification, recent works have demonstrated they are also extremely vulnerable to adversarial attacks. Most previous works have focused on attacking node classification networks under impractical white-box scenarios. In this work, we will propose a non-targeted Hard Label Black Box Node Injection Attack on Graph Neural Networks, which to the best of our knowledge, is the first of its kind. Under this setting, more real world tasks can be studied because our attack assumes no prior knowledge about (1): the model architecture of the GNN we are attacking; (2): the model's gradients; (3): the output logits of the target GNN model. Our attack is based on an existing edge perturbation attack, from which we restrict the optimization process to formulate a node injection attack. In the work, we will evaluate the performance of the attack using three datasets, COIL-DEL (\cite{riesen2008iam}), IMDB-BINARY (\cite{yanardag2015deep}), and NCI1 (\cite{wale2008comparison}). Our code and demo are publicly available for research purposes at: \url{https://github.com/bryanzhou008/Hard_Label_Black_Box_Attack_GNN.git} \footnote{This project was completed during the CS249: Graph Neural Networks class at UCLA advised by Prof. Yizhou Sun.}
\end{abstract}

\section{Introduction}
Nowadays, Graph Neural Networks (GNN) have been applied to various tasks, and demonstrated state-of-the-art performances. Among of the most popular tasks for Graph Neural Networks are node classification (\cite{kipf2016a}), link prediction (\cite{zhang2018link}), and graph classification (\cite{zhou2019meta, ying2018hierarchical, zhang2018link}). Specifically, our work focus on the graph classification tasks, in which a graph is given an associated label, for example, a group of actors might be labeled as "actors in Comedies," and the Graph Neural Network is trained to infer this label from the given graph. This kind of graph classification is especially useful in fields like social network applications, biomedics, chemistry, and object classification.

Previous studies also demonstrated that Graph Neural Networks are vulnerable to adversarial attacks (\cite{zugner2018adversarial, lin2020adversarial, dai2018adversarial}). However, the existing attack methods, as we summarized, have two main drawbacks. First, most attacking methods focus on node level prediction or edge level prediction, whereas only a few of them mentioned attack on graph level, which is actually a very important application of Graph Neural Networks. In a graph level attack, the adversarial agent will try to add/remove edge/node from the original graph to create adversarial instances to shift the model's prediction on the graph. Such attack can be very dangerous in real world scenarios. For example, when predicting the chemical properties of a compound in a medicine, then a wrong prediction can be deadly. And for user classification tasks in social network, malicious users may be able to change the label of the user group it is in and cause the model to give wrong recommendations to the users. Furthermore, existing adversarial attacks on Graph Neural Networks are usually non-practical in real-world scenarios. For example, there exist a portion of white-box attacks that assume adversarial agent has access to all necessary information about the target model, including its structure and gradients, and other attacks are soft-label black box attacks that assume agent has the output logits, all of these information is usually hidden from the public in real world. There are also some off-the-shelf hard-label black box attacks on Graph Neural Network, for example Jiaming et.al presented a hard label black box attack that perturb the edges in the graph to influence the graph level prediction (\cite{mu2021hard}). However, in applications like social network prediction, it is impossible for adversarial agents to modify connections between existing users. Therefore, in most cases, we believe node injection attack is more practical in the real-world and thus worth study more comparing to edge perturbing attacks. 

Therefore, in our paper, we propose a node injection attack on graph classification to better understand the robustness of Graph Neural Network in graph level tasks, and we assume a hard-label black box setting, meaning we only have access to the prediction label of the model, resembling most real world scenarios. And we evaluate the performance of our method on three datasets: COIL-DEL (100-class object classification) (\cite{riesen2008iam}), IMDB-BINARY(binary classification on social network) (\cite{yanardag2015deep}), and NCI1 (Binary classification of chemical property) (\cite{wale2008comparison}). For different datasets, we use different node feature initialization methods to generate nodes that can fit into the graph better (less noticable), and also provides a comparison between the performance of using different initialization methods. We further systematically evaluate the possible limitations and future works that can be built upon our work. To summarize, our main contribution in the work includes:
\begin{itemize}
    \item We present the first hard-label black box node injection attack on Graph Neural Networks, which we built upon an existing edge perturbing attack.
    \item We evaluate the performance of the method on three real world datasets, COIL-DEL (\cite{riesen2008iam}), IMDB-BINARY (\cite{yanardag2015deep}), and NCI1 (\cite{wale2008comparison}), using initialization methods specific to each dataset, and give a comparison between different initialization methods 
    \item We identify possible drawbacks or limitations of our current work and point out future research directions
\end{itemize}

\section{Related Work and Background}
\subsection{Adversarial Attack}
Depending on the availability of knowledge about the GNN model structure and output logits, we could categorize the attack into: White Box attack, where the adversarial agent has access to all information of the model, including gradient, model structure, and all outputs; Soft-Label Black Box Attack, where the adversarial agent only has access to the logits output of the model, and has no knowledge about the model’s gradient and structure; and the Hard-Label Black Box Attack, which is the attack we are deploying, only has access to the label output of the model.

\subsection{Adversarial Attack on GNN}
Multiple literature in the past have shown the vulnerability of Graph Neural Network against Adversarial Attack (\cite{zugner2018adversarial}). In general, there are two types of tasks that are frequently being conducted through GNN nowadays, Graph-level Classification and Node-level Classification, where we mainly focus on the former one in this paper. As for the forms of Adversarial Attack, which could be categorized into targeted attack and non-targeted attack. In our setting, we focused on the non-targeted attack, which means we tries to mislead to model to generate any prediction that deviates from the ground truth one instead of a specific prediction. Moreover, The attack can happen either during model training (poisoning) or during model testing (evasion)  in our case (\cite{ma2020towards}). For the specific method of adversarial attack, there are edge attack and node injection attack. To be specific, An attacker can perform an adversarial attack by perturbing one of three components in a graph: (i) perturbing nodes, i.e., adding new nodes or deleting existing nodes; (ii) perturbing node feature matrix, i.e., modifying nodes’ feature vectors; and (iii) perturbing edges, i.e., adding new edges, deleting existing edges or rewiring edges (\cite{mu2021hard}).

\section{Problem Formulation}

\begin{figure}[h]
\includegraphics[scale=0.375]{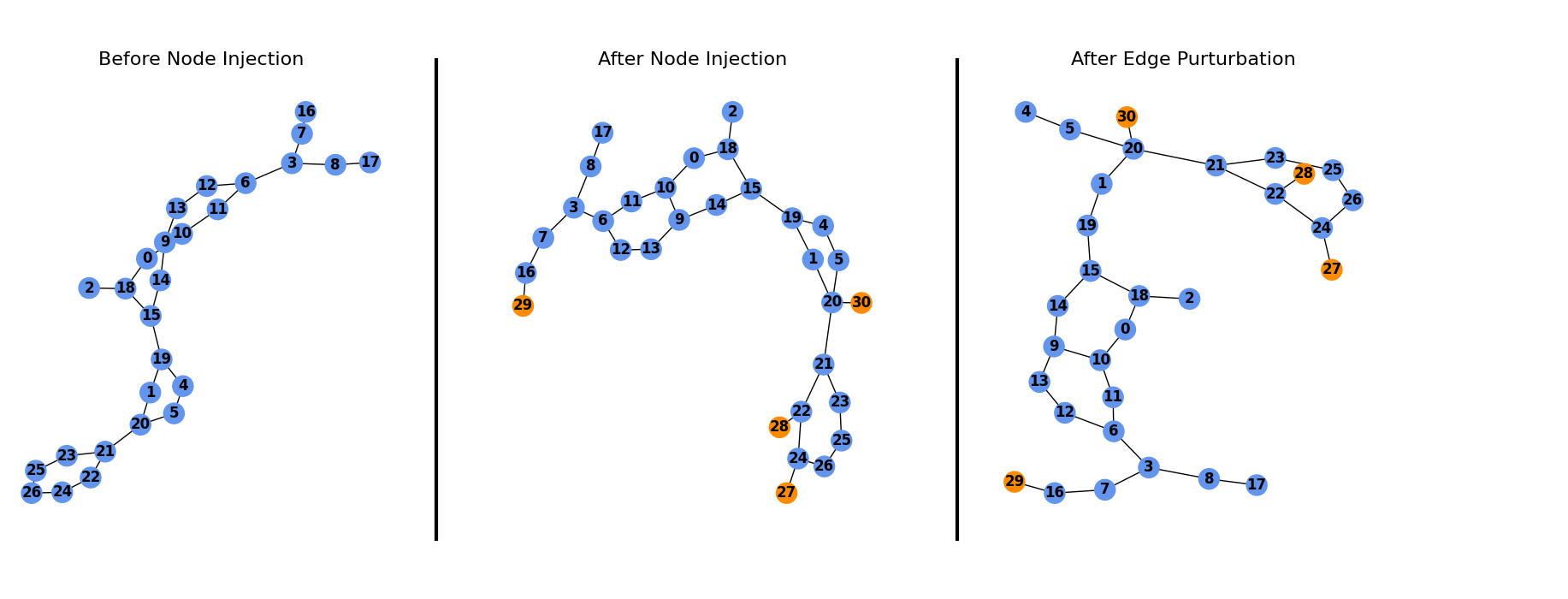}
\caption{A Set of Example Graphs to Illustrate Our Node Injection Attack Method}
\end{figure}

Given a target GNN model $f$ and a target victim graph $G$ with ground truth class prediction label $y_0$ for that particular input graph $G$, the attacker attempts to generate a non-targeted adversarial graph $G'$ by node injection with initialization on injected nodes' feature. Moreover, the attacker will initialize the connection of the $k$ injected nodes with $G$ to be the adjacency matrix $A$, and perturb $A$ to be $A'$ such that the predicted label of $G'$ will differ from $y_0$. As mentioned above in our setting, we only allows attacker to perturb the edges related to the $k$ injected nodes, but not the existing ones of the original graph, let the \textit{adversarial perturbation} be a binary matrix $\Theta \in \{0,1\}^{k \times N}$. Specifically, $\Theta_{ij} = 1$ means the attacker changes the edge status between nodes i and j and $\Theta_{ij} = 0$ means we remain the edge status between nodes i and j unchanged. Hence we can define the perturbation function $h$ that could generate the perturbed adjacency matrix $A'$ by $A' = h(A,\Theta)$ as:
\begin{align}
\begin{split}
    h(A,\Theta)_{ij} = 
    \begin{cases}
        A_{ij}          & \Theta_{ij} = 0\\
        \neg A_{ij}     & \Theta_{ij} = 1
    \end{cases}
\end{split}
\end{align}
Furthermore, we established a budget $b$ for to restrict the perturbation rate $r$ and hence we could construct the adversarial attack as the following optimization problem:
\begin{align}
\begin{split}
    \Theta^* = \underset{\Theta}{\arg\min} &\| A' - A \|_0, \\
    \text{subject to } &A' = h(A,\Theta), \\
    &f(A') \neq y_0, \\
    &r \leq b,
\end{split}
\end{align}
where r is defined as $r = \|A'-A\|_0/kN$ and $\|A'-A\|_0$ is the $L_0$ norm that counts the number of nonzero entries, which is the number of perturbation in our case.

However, we could show that the above optimization problem is intractable to solve due to $l_0$ norm and $\Theta$, and we need to reformulate the optimization problem by relaxing $\Theta$ to be continuous variables ranging from 0 to 1 that could be treated as the probability that corresponding edge between two nodes is changed, instead of binary entries {0,1}, such that we could approximate the gradients of the objective function. Therefore, the perturbation function $h$ could be reformulated as:
\begin{align}
\begin{split}
    h(A,\Theta)_{ij} = 
    \begin{cases}
        A_{ij}          & \Theta_{ij} < 0.5\\
        \neg A_{ij}     & \Theta_{ij} \geq 0.5
    \end{cases}
\end{split}
\end{align}
We can then define a new objective function that replaces the $L_0$ norm with the $L_1$ norm. Define distance function $g(\Theta)$ that measures the distance from the target graph to the classification boundary as:
\begin{align}
    g(\Theta) = \underset{\lambda > 0}{\arg\min} \{f(h(A,\lambda\Theta_{norm}) \neq y_0\},
\end{align}
where $\Theta_{norm}$ is the normalized perturbation matrix such that $\|\Theta_{norm}\|_2 = 1$. Therefore, $g(\Theta)$ gives the minimum distance $\lambda$ starting from $A$ towards another class in direction $\Theta$ that changes the prediction label. In addition, we denote $\hat{g}(\Theta) = g(\Theta)\Theta_{norm}$ to be the distance vector.

Although it may seem reasonable to get optimal $\Theta^*$ by simply minimize $g(\Theta)$, it is not effective since it does not take into account of the search direction. Two perturbations with equal distance but different $\Theta$ might result in a different number of perturbations on edges (\cite{mu2021hard}). Hence, we define the following objective function: $p(\Theta) = \|clip(\hat{g}(\Theta)-0.5)\|_0$, where $clip(x)$ clips $x$ into $[0,1]$, and this objective function counts the number of elements of $\hat{g}(\Theta)$ that exceed 0.5. And to make the calculation of gradients viable, we replace the $L_0$ norm with the $L_1$ norm as:
\begin{align}
    p(\Theta) = \|clip(\hat{g}(\Theta)-0.5)\|_1.
\end{align}
Finally, we could find the optimal matrix $\Theta^*$ by minimizing $p(\Theta)$ and convert the original optimization problem in Eq.(2) into:
\begin{align}
\begin{split}
    \Theta^* &= \underset{\Theta}{\arg\min} \text{\space} p(\Theta), \\ &\text{subject to } r \leq b.
\end{split}
\end{align}

\section{Experiments}

For experiments, our target model will be a Graph Isomorphism Network (GIN) model trained on COIL-DEL (\cite{riesen2008iam}), IMDB-BINARY (\cite{yanardag2015deep}), or NCI1 (\cite{wale2008comparison}). A summary of these three datasets is listed in table [1]. For these three datasets, in order to make the injected node less noticeable to model or defender, we will use three different node feature initialization methods that are designed specifically to each dataset. Moreover, we will compare the effect of using different node connection initialization methods, namely randomly connect injected node to an existing node (random initialization), or connect the injected node to the node with highest degree (mode initialization). Theoretically we expect a higher success rate when using mode initialization. Note that in IMDB-BINARY dataset, we use an iterative approach because their node feature initialization methods are deterministic (non-random). We will first try to inject 1 node, and if the attack fails, we try to inject 2, so on. Until we exhaust a preset budget on the number of nodes we can inject or we succeeded in an attack.

\begin{table}[h]
\centering
\begin{tabular}{||c | c | c | c||} 
 \hline
 Dataset & COIL & IMDB-B & NCI1 \\ [0.5ex] 
 \hline\hline
 Num. of Graphs & 390 & 100 & 411 \\ 
 \hline
 Num. of Classes & 100 & 2 & 2 \\
 \hline
 Avg. Num. of Nodes & 19.77 & 21.54 & 29.87 \\
 \hline
 Avg. Num. of Edges & 96.53 & 54.24 & 32.30 \\
 \hline
\end{tabular}
\caption{Summary of datasets}
\centering
\end{table}

The metrics in the tables we present the experiment results are as follow:
\begin{itemize}
    \item method: the method for connection initialization
    \item budget: the maximum percentage of nodes allowed to inject
    \item SR: success rate (percentage) of the attack, = (success + Pred Change) / (Num. Graphs - No need)
    \item success: number of instances that have prediction changed after the direction search and binary search
    \item Pred Change: number of instances that have prediction changed right after initialization, in experiments we consider this as successful attack as well, because theoretically we will be able to change the label after performing the algorithm if the label is already changed after initialization
    \item Injected: the percentage of instances that have adversarial nodes still connected to the graph after perturbing the graph
    \item No need: the number of instances that the GIN model originally predicted wrong, thus no meaning to attack
    \item Perturb Edge: average number of edges flipped in the attack of the whole dataset
\end{itemize}

\subsection{COIL-DEL}
In the COIL-DEL dataset, the node features are $(x,y)$, $x,y \in \mathbb{N}$ pairs representing the coordinate of the node in the space. Therefore, when initializing the feature of the node to be injected, denote the nodes in the original graph as $(x_i,y_i)$ for $i \in [0,n-1]$, where n stands for the number of nodes in the graph, we generate random pair $(x,y)$ with $x \sim \mathcal{N}(mean(x_0,...x_n),std(x_0,...x_n))$, $y \sim \mathcal{N}(mean(y_0,...y_n),std(y_0,...y_n))$, and round $(x, y)$ to the nearest integer. The experiment results on COIL-DEL is shown in table [2]. From the table, we can observe that when using mode connection initialization, we indeed get higher success rate at both 10\% and 15\% node injection budget as we expected. Moreover, if we look at the Pred Change column, we can see that when using random connection initialization, Pred Change decreased a lot, which makes sense because we are expecting less noticeable changes after initialization when we use random connection. Moreover, the experiments demonstrated that just initializing the connection is sufficient for keeping the new nodes connected to the graph through out the attack, without any further restriction during the attacking process.

\begin{table}[H]
\centering
\begin{tabular}{||c | c | c | c | c | c | c | c||} 
 \hline
 method & budget & SR & success & Pred Change & Injected & No need & Perturb Edge \\ 
 \hline\hline
 mode & 0.1 & \textbf{50.35} & 94 & 51 & 99.16 & 102 & 13.51 \\ 
 \hline
 mode & 0.15 & \textbf{63.89} & 112 & 72 & 100 & 102 & 11.57 \\
 \hline
 random & 0.1 & 48.68 & 100 & \textbf{40} & 100 & 102 & 15.51 \\
 \hline
 random & 0.15 & 61.46 & 122 & \textbf{55} & 100 & 102 & 17.38 \\
 \hline
\end{tabular}
\caption{Experiment Results for COIL-DEL dataset}
\centering
\end{table}

\subsection{IMDB-BINARY}

For the IMDB-BINARY dataset, we observed that the node feature are just all one's in the original graph, so node feature initialization is very trivial in this case. In table [3] we show the experiment results on this dataset. For Success Rate and Pred Change, we observe similar trends as in COIL-DEL dataset.

\begin{table}[H]
\centering
\begin{tabular}{||c | c | c | c | c | c | c | c||} 
 \hline
 method & budget & SR & success & Pred Change & Injected & No need & Perturb Edge \\ 
 \hline\hline
 mode & 0.1 & \textbf{27.63} & 7 & 14 & 100 & 24 & 167.15 \\ 
 \hline
 mode & 0.15 & \textbf{53.95} & 20 & 21 & 100 & 24 & 104.4 \\
 \hline
 random & 0.1 & \textbf{27.63} & 6 & \textbf{15} & 100 & 24 & 191.67 \\
 \hline
 random & 0.15 & 51.32 & 20 & \textbf{19} & 100 & 24 & 111.58 \\
 \hline
\end{tabular}
\caption{Experiment Results for IMDB-BINARY dataset}
\centering
\end{table}

\subsection{NCI1}

For the NCI1 dataset, the node features are 37-dimensional one-hot encodings for the atom type that the node represents. We randomly generate injected node feature. For atom type x, we generate it with probability $\frac{count(x)}{Num.Nodes}$. For SR and Pred Change, we observe similar trends as the former 2 experiments.

\begin{table}[H]
\centering
\begin{tabular}{||c | c | c | c | c | c | c | c||} 
 \hline
 method & budget & SR & success & Pred Change & Injected & No need & Perturb Edge \\ 
 \hline\hline
 mode & 0.1 & \textbf{47.43} & 101 & 45 & 99.17 & 104 & 128.17 \\ 
 \hline
 mode & 0.15 & \textbf{63.19} & 140 & 21 & 98.82 & 104 & 165.06 \\
 \hline
 random & 0.1 & 41.04 & 101 & \textbf{25} & 100 & 104 & 143.63 \\
 \hline
 random & 0.15 & 54.40 & 140 & \textbf{27} & 100 & 104 & 185.53 \\
 \hline
\end{tabular}
\caption{Experiment Results for IMDB-BINARY dataset}
\centering
\end{table}

\begin{figure}[h]
\includegraphics[scale=0.356]{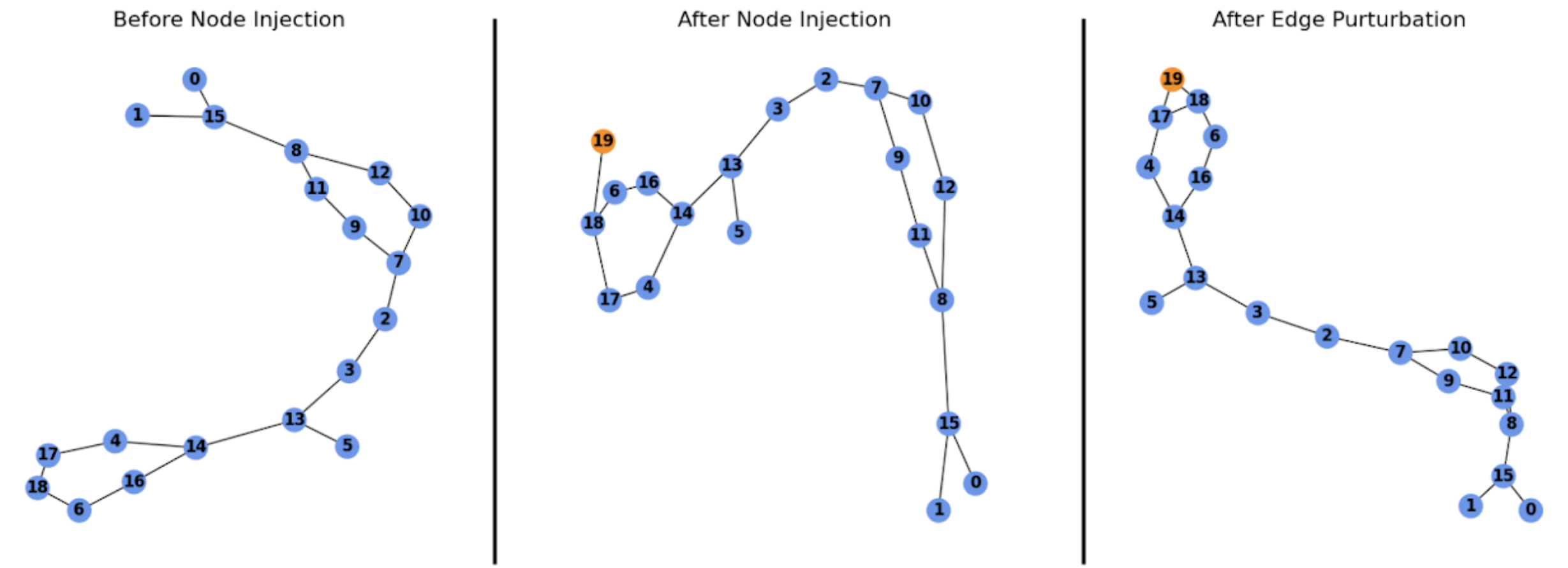}
\caption{An Example Graph From the NCI1 Dataset Successfully Attacked by Our Model}
\end{figure}

\subsection{Further Observersation for IMDB-BINARY and NCI1 Dataset}

In our experiments for IMDB-BINARY and NCI1 datasets, we can see from the results that the Perturb Edge in these 2 sets of experiments are significantly larger than that of the COIL-DEL dataset (10-20 VS 100+). We believe this is because COIL-DEL dataset has 100 class labels, whereas IMDB-BINARY and NCI1 are binary classification datasets. As a result, if we only search for perturb direction restricted on the injected nodes, we may need to go a larger distance along the direction we find to arrive at the decision boundary comparing to in the COIL-DEL dataset, leading to a much larger Perturbation number. If we compare the Perturb Edge stats to the average number of edges in IMDB-BINARY and NCI1 datasets, we will notice that these two sets of numbers are somewhat similar. Therefore, we believe our method still needs improvements on such datasets where the decision boundary can be very far away from the original instance. For example, future research may seek to define a better direction search algorithm to find a direction along which we walk to decision boundary in a smaller distance. Alternatively, one may consider a better connection initialization method to allow to adversarial nodes to have larger impact on the graph classification.

\section{Extended investigation on feature of injected nodes}
\subsection{Methodology}
In all above investigations, we assume injected node features either to be random or to be the same as the mean of all nodes in current graph. However, such assumption does not take fully advantage of conducting attacks on graph using node injections. We ignore the features of the injected nodes. Based on intuition, if we can manipulate the features of the injected nodes to make it has the biggest impact on the graph classification task, we can definitely make attacks more successful and more efficient. 

Since it is hard to conduct optimization process in a hard-label black box attack setting, we only focus on manipulating the injected features at the initialization phase. In a graph classification task, due to the fact that graph neural network will aggregate information from its neighbors, we believe the node with highest rank in the graph should have the greatest influence on the final prediction of the graph, and we can such node as pivot node. So we try to interfere this aggregation process by injecting nodes that have very different features from the features of the pivot node and connect all injecting nodes to the pivot node at initialization. In addition, to make sure the features of the injected nodes make sense in all datasets and also make the injection less noticeable. We choose features from the original graph and slightly perturb the feature to make sure it is not same to any node in the graph or to any other injected nodes. Since the selected features come from original graph, the feature itself always makes sense in all datasets. Also, since the features are not going to same as any other features in the graph, we make the injection process less noticeable.

\subsection{Experiments and analysis}
For experiments on our manipulation of the injected nodes features, we decide to only run the experiments on COIL-DEL dataset with above algorithm. We run the both iterative version where we gradually increase the number of nodes injected until certain budget and the normal version where we directly injected fixed percentage of the nodes to the original graph. The following is our final result. 

\begin{table}[h]
\centering
\begin{tabular}{||c | c | c | c | c | c | c | c||} 
 \hline
 method & budget & SR & Perturb Edge & Attack Time & Query Count\\ 
 \hline\hline
 Normal & 0.1 & 51.61 & 10.84 & 9.06 & 544.2 \\ 
 \hline
 Iterative & 0.1 & 51.95 & 8.25 & 11.1 & 580.8 \\
 \hline
 Normal & 0.15 & 64.74 & 10.65 & 9.56 & 644.0 \\
 \hline
 Iterative & 0.15 & 65.17 & 8.65 & 13.2 & 786.6 \\
 \hline
 Normal & 0.2 & 71.18 & 11.88 & 9.74 & 632.0 \\
 \hline
 Iterative & 0.2 & 71.53 & 9.39 & 15.7 & 822.8 \\
 \hline
\end{tabular}
\caption{Feature manipulation experiment result on COIL-DEL dataset}
\centering
\end{table}

As you can see, with feature manipulation using mode connection, we can achieve a slightly higher success rate with a lower number of edge perturbations. We prove this is indeed a smarter initialization that makes the attack more successful and less noticeable with less changes to the original graph. 

\section{Conclusion and Future Works}
We believe there are three main contributions in our work. Firstly, we extend on the original work to significantly improves its success rate with single-node injection and perturbations to edges in the whole graph. Secondly, we propose method to do node injections with random or mean feature initialization work in a strict black box setting. To our knowledge, we are the first who investigate in this area. Thirdly, we propose smarter ways to initialize the feature of the injected nodes and we prove it does lead to a higher success rate with smaller changes to the original graph. 

There are definitely still a lot of room for further investigation based on our current work. Most importantly, since currently our injected features are all fixed after initialization, how to develop a concrete method to update and find the optimal injected node feature is a very interesting topic. Some other future works include finding a better initial search specific to node injections as the original edge perturbation paper, improve our performance on datasets that have small number of classes, and find better connection initialization for injected nodes that is less noticeable. We hope our work can inspire further researches in related fields.

\printbibliography[title={Reference}]

\section{APPENDIX}
\subsection{Interactive Demo}
We have created an interactive Google Colab Demo of our 
\href{https://colab.research.google.com/drive/1S4hpKZexG_wr3CHhUJx3ZgXewJ5XXCUn#scrollTo=duDaB7pxdBFg}{Node Injection Attack}. When using the demo, you should first change google drive paths to your own and run all cells collapsed under the "Environment" group to install all dependencies. Then, you may continue to the "Settings" block to set all parameters you wish to try. Here dataset name should be one of 'NCI1', 'COIL-DEL', and 'IMDB-BINARY'. Injection number specifies how many nodes you want to inject to each graph. Injection percentage specifies what percentage of total number of nodes in the graph you wish to inject. Only one field of the above two should be non-zero. Initialization specifies how the node features will be initialized. The supported methods include zero, one, random, and node\_mean(the mean value of all node features in the graph). Connection specifies how and whether you want the injected nodes to be connected to the graph. The available options include no\_connection, random and mode(connect all injected nodes to the node in the original graph with most connections).

Once you have specified all the parameters in "Settings", please run the blocks collapsed in "Utility Functions" to read in data and. Now, to visualize the injection results, you should just run the first cell in "Demo" and the injection results should show up on your screen.

\begin{figure}[h]
\includegraphics[scale=0.5073]{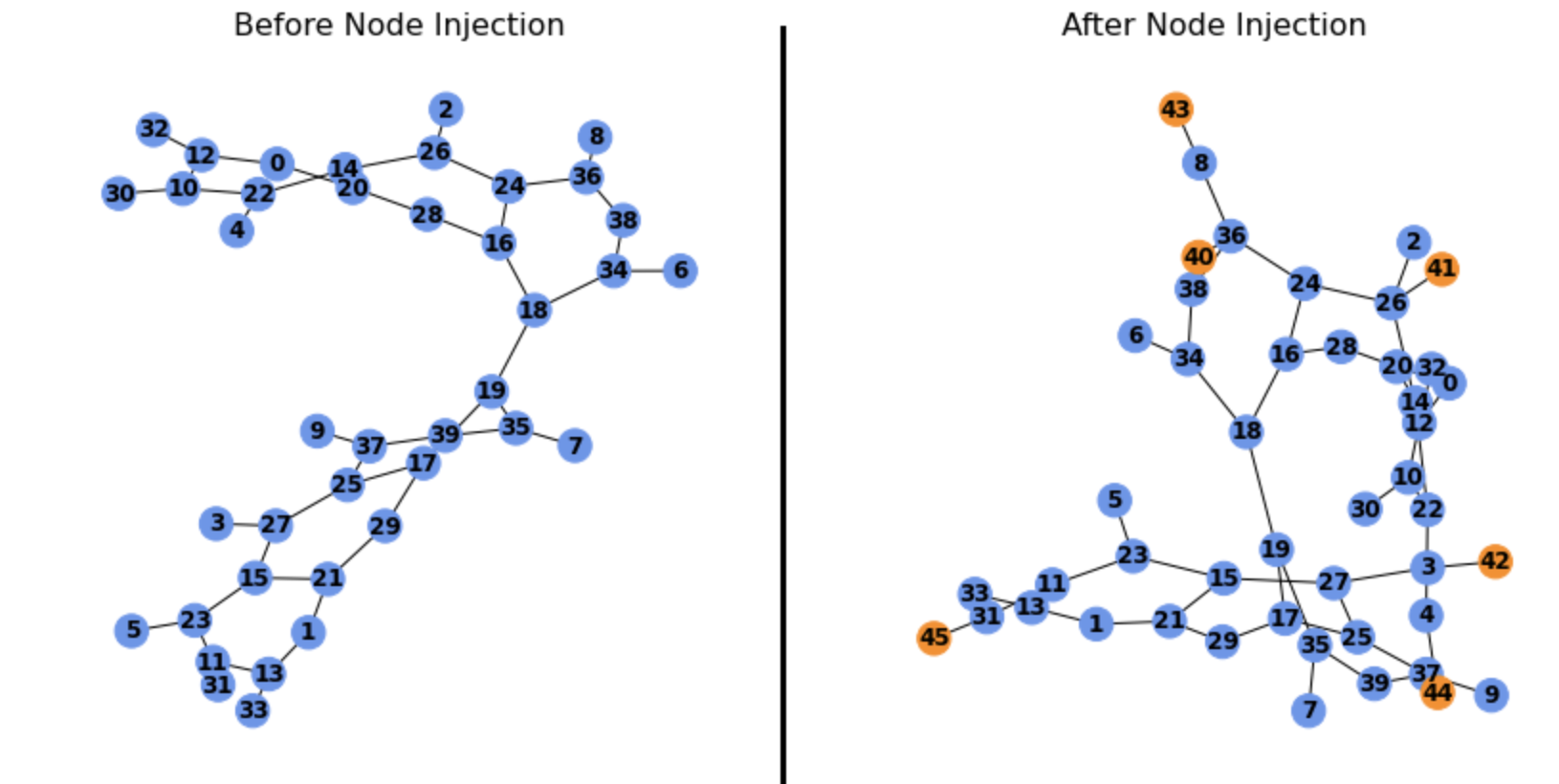}
\caption{An example graph produced from the following settings: \{dataset: NCI1, injection number: 0, injection percentage: 0.15, initialization method: node mean, connection method: random\}}
\end{figure}

\end{document}